\pdfoutput=1

\documentclass[11pt]{article}
\usepackage{algorithm}
\usepackage{algpseudocode}
\usepackage{adjustbox}
\usepackage[]{acl}
\usepackage[inline]{enumitem}
\usepackage{times}
\usepackage{latexsym}
\usepackage{booktabs} 
\usepackage{amsmath} 
\usepackage[T1]{fontenc}
\usepackage{multirow}

\usepackage[utf8]{inputenc}

\usepackage{microtype}

\usepackage{inconsolata}

\AtBeginDocument{%
  \providecommand\BibTeX{{%
    \normalfont B\kern-0.5em{\scshape i\kern-0.25em b}\kern-0.8em\TeX}}}
            
\title{Unlocking the Power of Large Language Models for Entity Alignment}

\author{Xuhui Jiang$^{1,2,3}$, Yinghan Shen$^{1}$ , Zhichao Shi$^{1,2}$ , Chengjin Xu$^{3}$ , \textbf{Wei Li}$^{1}$ , \textbf{Zixuan Li}$^{1}$ , \\ \textbf{Jian Guo}$^{3}$ , \textbf{Huawei Shen}$^{1}$ , \textbf{Yuanzhuo Wang}$^{1}$ \\
  1. CAS Key Laboratory of AI Safety, Institute of Computing Technology, CAS \\
  2. School of Computer Science and Technology, University of Chinese Academy of Science \\
  3. IDEA Research, International Digital Economy Academy \\
  \texttt{\{jiangxuhui, xuchengjin, guojian\}@idea.edu.cn} \\
  \texttt{\{shenyinghan,  shizhichao, liwei2023, lizixuan, wangyuanzhuo\}@ict.ac.cn}}

\begin{document}
\maketitle

\begin{abstract}
Entity Alignment (EA) is vital for integrating diverse knowledge graph (KG) data, playing a crucial role in data-driven AI applications.
Traditional EA methods primarily rely on comparing entity embeddings, but their effectiveness is constrained by the limited input KG data and the capabilities of the representation learning techniques.
Against this backdrop, we introduce ChatEA, an innovative framework that incorporates large language models (LLMs) to improve EA.
To address the constraints of limited input KG data, ChatEA introduces a KG-code translation module that translates KG structures into a format understandable by LLMs, thereby allowing LLMs to utilize their extensive background knowledge to improve EA accuracy.
To overcome the over-reliance on entity embedding comparisons, ChatEA implements a two-stage EA strategy that capitalizes on LLMs’ capability for multi-step reasoning in a dialogue format, thereby enhancing accuracy while preserving efficiency.
Our experimental results verify ChatEA's superior performance, highlighting LLMs' potential in facilitating EA tasks.
The source code is available at \href{https://github.com/IDEA-FinAI/ChatEA}{https://github.com/IDEA-FinAI/ChatEA}.

\end{abstract}

\section{Introduction}

Entity alignment (EA) aims at aligning entities from diverse knowledge graphs (KGs).
It is a pivotal step in unifying data from heterogeneous sources and plays a crucial role in data-driven AI.
Current EA methods predominantly rely on measuring the similarity of entity embeddings derived from knowledge representation learning (KRL) techniques. These techniques learn from the topology and semantics of KGs to derive entity embeddings~\cite{zhang2022benchmark}. 
However, these methods fail to incorporate the external knowledge of entities, which is essential for the alignment process.
Moreover, the KRL-based alignment methods merely calculate the similarity between two entity embeddings in a black-box manner, which lacks a detailed and explicit reasoning process for alignment.
Such limitations significantly affect the performance of EA methods, especially in aligning highly heterogeneous KG pairs where KRL-based methods struggle to capture the complex correlations among KGs~\cite{jiang2023rethinking}. 

\begin{figure}[t]
    \centering
    \includegraphics[width=1\columnwidth]{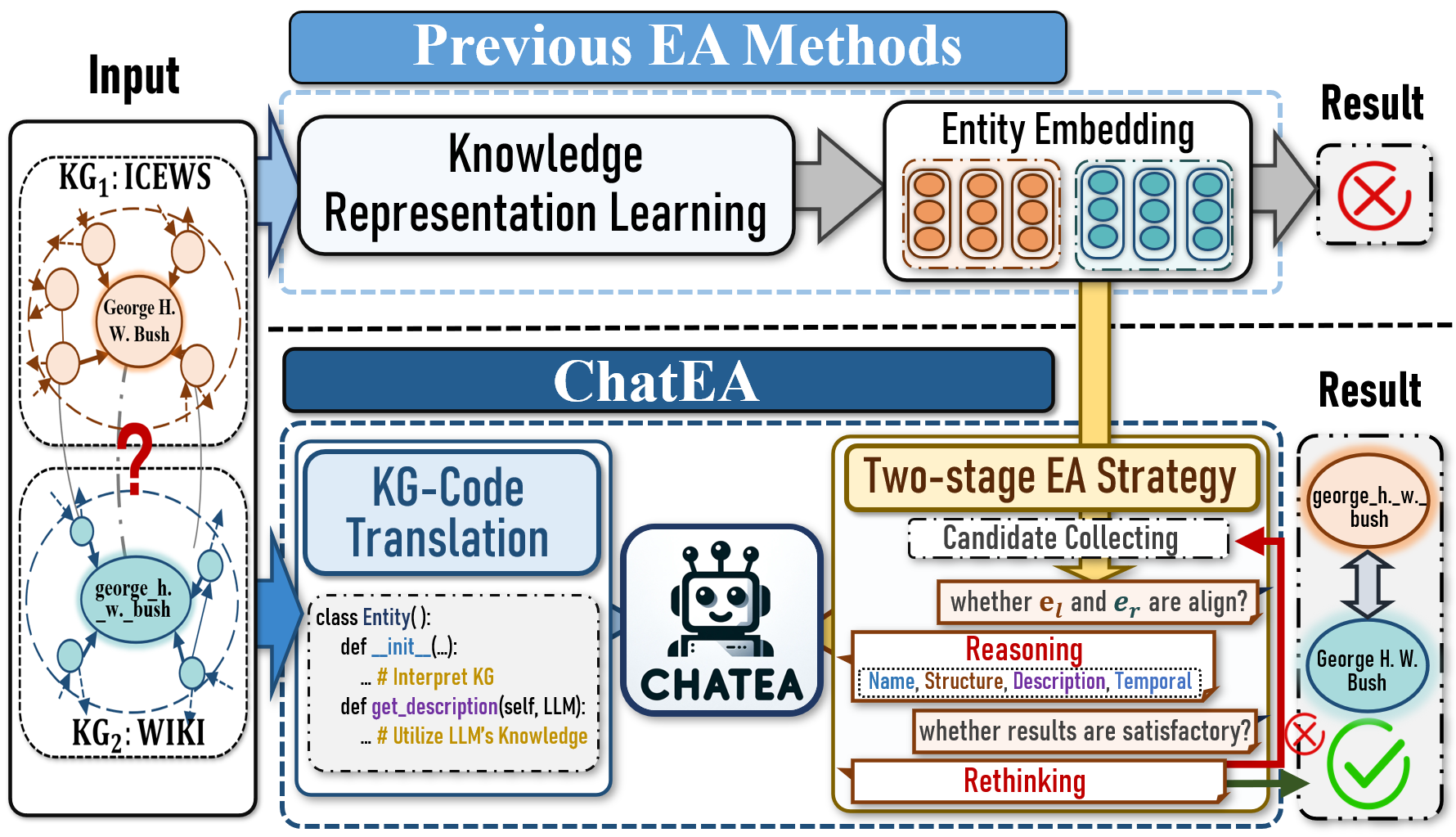}
    \caption{A comparison of previous EA and ChatEA.}
    \label{fig:fig_1}
    \vspace{-15pt}
\end{figure}

Recently, large language models (LLMs) have showcased their effectiveness across a range of natural language processing tasks, revealing a vast but under-explored potential in EA.
These LLMs are trained on extensive corpora to encapsulate external knowledge, offering a rich source of contextual information for entities in KGs~\cite{jiang2023evolution}.
Furthermore, the recent studies of adopting LLMs in knowledge extraction and reasoning also demonstrated their strong reasoning abilities on KGs~\cite{gui2023instructie, wei2023zero}.
These features of LLMs offer a promising path to overcome the constraints faced by current EA methods.


In this paper, we propose ChatEA, a novel framework designed to enhance KRL-based EA methods by utilizing the extensive background knowledge and reasoning abilities of LLMs.
As shown in Figure~\ref{fig:fig_1}, ChatEA integrates KRL-based EA methods in the feature pre-processing phase to assist LLMs in the subsequent selection of candidate entities.
To overcome the constraints of limited input KG data, ChatEA firstly features a KG-Code translation module. The module initially converts KGs into a code format, explicitly accounting for entity definition toward LLMs' comprehension of graph-structured KG data~\cite{yang2024if, li2024knowcoder}. 
Then it facilitates the generation of entity descriptions leveraging LLMs' background knowledge.
To overcome the over-reliance on comparing entity embeddings for EA and improve transparency, ChatEA employs a two-stage EA strategy, leveraging LLMs’ multi-step reasoning in dialogue form to enhance accuracy and maintain efficiency.
During the candidate collecting stage, ChatEA identifies potential entities by comparing embeddings derived from the earlier feature pre-processing phase.
In the reasoning and rethinking stage, it evaluates the likelihood of alignment between entity pairs by comprehensively considering the name, structure, entity description, and temporal information, and then decides whether to broaden the search scope and continue subsequent iterations.

We validated our method on two conventional EA datasets: DBP15K(EN-FR) and DBP-WIKI, along with two more challenging but practical datasets: ICEWS-WIKI and ICEWS-YAGO, characterized by their highly heterogeneous KGs and the complexity of capturing inter-KG correlations~\cite{jiang2023rethinking}. The extensive experiments reveal ChatEA's superiority over existing state-of-the-art EA methods and underscore the potential of LLMs in enhancing EA performance. Notably, ChatEA significantly improves Hits@1 by 9\%-16\% compared to previous state-of-the-art methods on the two challenging datasets.

In general, our main contributions are as follows:

\indent (1) To solve the limitations of the existing KRL-based EA methods, we explore the potential of adopting LLMs for better EA performance.

\indent (2) We design ChatEA, a novel framework that integrates LLMs with KRL-based EA methods for enhanced EA performance.

\indent (3) We conduct extensive experiments to evaluate the effectiveness of ChatEA, and discuss the value and limitations of LLMs in EA tasks.

\begin{figure*}[t]
    \centering
    \includegraphics[width=2\columnwidth]{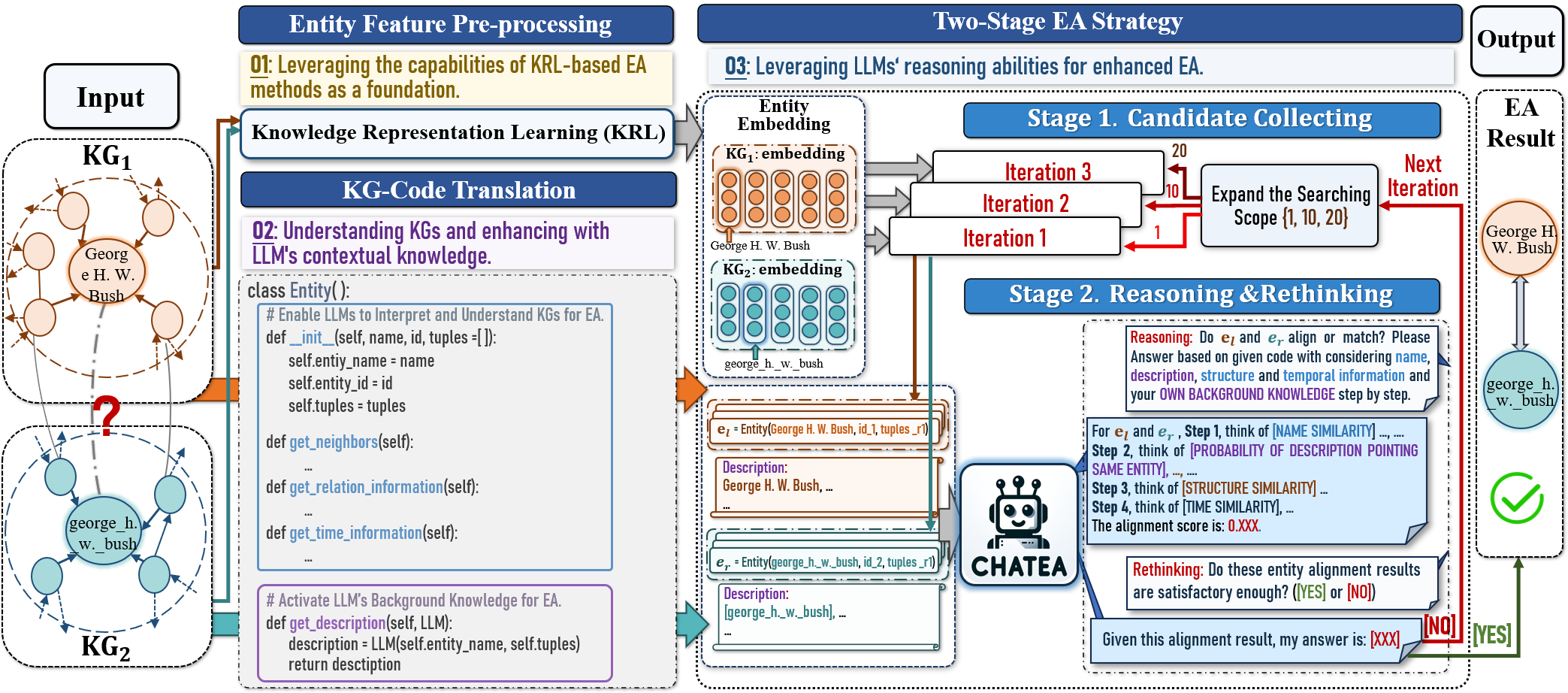}
    \caption{The overview framework of our proposed ChatEA, which consists of three main components: (1) entity feature pre-processing, (2) KG-code translation, and (3) two-stage EA strategy.}
    \label{fig:fig_2}
    \vspace{-10pt}
\end{figure*}

\section{Preliminaries and Related Works}
This section first delineates the relevant definitions, followed by an overview of related works.

\subsection{Preliminaries}
\textbf{Knowledge graph (KG)} $\mathcal{KG} = (\mathcal{E}, \mathcal{R}, \mathcal{F})$ stores the real-world knowledge in the form of facts $\mathcal{F}$, given a set of entities $\mathcal{E}$ and relations $\mathcal{R}$, the $ {(e_{head}, r, e_{tail})} \in \mathcal{E}\times \mathcal{R} \times \mathcal{E}$ denotes the set of facts $\mathcal{F}$, where $e_{head}, e_{tail} \in \mathcal{E}$ respectively denote the head entity and tail entity, $r \in \mathcal{R}$ denotes the relation.
For the temporal information in KGs, given timestamps $\mathcal{T}$, we denote $t \in \mathcal{T}$ as the temporal information of the facts, and each fact is represented in the form of ${(e_{head}, r, e_{tail},t)}$.

\textbf{Entity alignment (EA)} is a crucial task in KG research. Given two KGs, $\mathcal{KG}_1=(\mathcal{E}_1, \mathcal{R}_1, \mathcal{F}_1)$ and $\mathcal{KG}_2=(\mathcal{E}_2, \mathcal{R}_2, \mathcal{F}_2)$, the goal is to determine the identical entity set $\mathcal{S} = {(e_i, e_j)|e_i\in \mathcal{E}_1, e_j \in \mathcal{E}_2}$. In this set, each pair $(e_i, e_j)$ represents the same real-world entity but exists in different KGs.

\subsection{Related Works}
Within the EA domain, various knowledge representation learning methods have been mainly adopted, generally categorized into three types: translation-based, GNN-based, and other methods.

Translation-based methods, like MTransE~\cite{chen2017multilingual}, BootEA~\cite{sun2018bootstrapping}, and AlignE~\cite{sun2018bootstrapping}, founded on TransE's framework~\cite{bordes2013translating}, excel in knowledge representations.
Graph Neural Networks (GNNs), exemplified by GCN~\cite{kipf2016semi}, mark a notable advance in EA by aggregating neighborhood information to generate entity embeddings. GCN-Align~\cite{wang2018cross}, RDGCN~\cite{wu2019relation}, AttrGNN~\cite{AttrGNN}, and Dual-AMN~\cite{mao2021boosting} exemplify GNN-based EA methods, utilizing GCN for modeling structure information and learn entity embedding. Recent GNN-based methodologies, e.g., TEA-GNN~\cite{xu2021time}, TREA~\cite{xu2022time}, and STEA~\cite{STEA}, have integrated temporal data, underscoring its significance in EA.
Other approaches, such as BERT-INT~\cite{tang2020bert} and TEA~\cite{TEA} employ language models to improve the expressiveness of entity embeddings for EA. Fualign~\cite{wang2023fualign}, SDEA~\cite{SDEA}, and Simple-HHEA~\cite{jiang2023rethinking} address the heterogeneity in KGs by utilizing side information.


The above three types of EA methods, while widely adopted, exhibit significant limitations. They often heavily rely on the quality of input KG data, and most of their successes are dependent on the quality of entity embeddings.
This dependence poses challenges in scenarios where KGs are highly heterogeneous or when the quality of the embeddings is suboptimal~\cite{jiang2023rethinking}.

In light of these limitations, there emerges an urgent need to explore new paradigms for EA tasks.
LLMs, with their extensive parametric knowledge, offer a compelling alternative. Their ability to process KGs without the sole reliance on representation learning positions them as a promising solution to the limitation of previous methods.

\section{Method}
In this section, we present ChatEA, a novel framework that unlocks the power of LLMs for EA tasks. We meticulously design the architecture of ChatEA around three pivotal objectives:



\vspace{-4pt}
\begin{itemize}[leftmargin=*]
    \item \textbf{O1: Leveraging the capabilities of KRL-based EA methods as a foundation:} This objective aims to utilize the strengths of KRL-based EA methods while circumventing its over-reliance on entity embedding similarity comparisons.
    \vspace{-4pt}
    
    \item \textbf{O2: Understanding KGs and enhancing with the external knowledge in LLM:} This objective aims to enable LLMs to comprehend KGs effectively, and activate LLMs' extensive background knowledge to enrich KGs, thus providing a more nuanced foundation for EA tasks.
    \vspace{-4pt}

    \textbf{O3: Leveraging LLMs' reasoning abilities for enhanced EA:} It aims to utilize the reasoning prowess of LLMs in EA, improving accuracy and transparency while balancing efficiency.
    \vspace{-2pt}
\end{itemize}

\subsection{Overview of the ChatEA Framework}
As illustrated in Figure~\ref{fig:fig_2}, the architecture of ChatEA is designed to enhance EA by integrating the background knowledge and reasoning capabilities of LLMs with the basic strengths of KRL.

In response to the objective \textbf{O1}, ChatEA initially leverages KRL-based EA techniques to assimilate entity features, such as names, structural, and temporal attributes, into embeddings, which assist LLMs in the candidate entity selection.

In response to the objective \textbf{O2}, ChatEA's KG-Code translation module plays a critical role.
This module transforms the KG into a coded format through class initialization and function, then adopts LLMs for description generation, thus bridging the KGs with LLM's background knowledge.

In response to the objective \textbf{O3}, ChatEA introduces a two-stage EA strategy.
This involves pre-processed entity embeddings to swiftly collect candidate entities, then engages LLMs for iterative reasoning and rethinking alignment probabilities of the target and candidate entities in a dialogue form.

\subsection{Entity Feature Pre-processing}~\label{SEC_feature_pre_processing}
In response to objective \textbf{O1}, we leverage the capabilities of knowledge representation learning to model entity information for EA by adopting the Simple-HHEA~\cite{jiang2023rethinking}, noted for its simplicity and effectiveness in generating entity representations.
Initially, we employ BERT~\cite{devlin2018bert} to derive semantic embeddings of entity names, followed by dimension reduction via feature whitening transformation~\cite{su2021whitening}. Temporal attributes are represented using Time2Vec~\cite{kazemi2019time2vec}, converting time into a learnable vector. Additionally, we incorporate structural information through a biased random walk method~\cite{wang2023fualign}, which optimally balances BFS and DFS techniques for precise one-hop and multi-hop relational analysis. The culmination of these processes results in final embeddings that merge name, temporal, and structural features into a unified multi-view representation for each entity. This multi-view preprocessing strategy is refined by Margin Ranking loss for training and Cross-domain Similarity Local Scaling (CSLS)~\cite{conneau2017word} for similarity measurement, aiding LLMs in the subsequent selection of candidate entities. The detailed feature pre-processing pipeline can be found in Appendix~\ref{appendix_feature}.

Additionally, ChatEA's plug-and-play characteristic enables it to integrate with various KRL-based EA methods as a foundation, which influences the quality of embeddings used for entity similarity comparison. We also conduct additional ablation experiments in Section~\ref{SEC_entity_embedding} to assess their impact on ChatEA's performance.

\subsection{KG Input and Understanding in LLMs}
In the ChatEA framework, the KG-Code translation module stands as a pivotal solution for the second objective \textbf{O2}. The module inputs and understands KGs in LLMs and activates the LLMs' extensive background knowledge for EA.

\subsubsection{Understanding Knowledge Graphs}
The efficacy of the code format in aiding the LLMs to process graph-structured KG data has been demonstrated in prior research~\cite{yang2024if}.
In light of this, we propose the KG-Code translation module to describe entity information, which is defined with a Python-style class, and comprised of five member functions. These functions are specifically designed to convert entity attributes into a distinct data structure and subsequent visitation, thereby facilitating a more comprehensive understanding by the LLM. 
Specifically, as shown in Figure~\ref{fig:fig_2}, The \texttt{\_\_init\_\_()} function enables LLMs to process entity attribute information initially.
Given an entity, the \texttt{get\_neighbors()}, \texttt{get\_relations()}, and \texttt{get\_temporal()} member functions enable LLMs to understand neighborhoods, relations, and temporal information about entities contained in KGs' tuples.

\subsubsection{Activating LLM’s Inherent Knowledge}
The \texttt{get\_description()} function of the KG-Code translation module also addresses the activation of LLM's inherent knowledge in the context of EA.
The prompt is designed to encourage the LLM to autonomously produce concise descriptions of entities. These descriptions are subsequently utilized in the EA procedure. Subsequently, the system prompt, along with few-shot examples and prompts, emphasizes the use of the LLM's background knowledge in reasoning procedure.

By employing these strategies, the KG-Code translation module integrated into ChatEA not only facilitates the processing and comprehension of KG data by LLMs but also leverages their background knowledge for effective EA.

\subsection{Two-Stage EA Strategy in ChatEA}
In enhancing the accuracy while balancing efficiency in objective \textbf{O3}, we propose a two-stage EA strategy.
The candidate collecting stage selects candidate entities via similarity comparison grounded in pre-processed entity representation. Subsequently, it reevaluates whether these results necessitate an expansion of the search parameters for subsequent iterations.

\subsubsection{Stage 1: Candidate Collecting}
This stage leverages entity embeddings obtained from pre-processing to filter out candidate entities.
The process begins with identifying the most probable candidates, for a given target entity, ChatEA utilizes the entity embeddings derived from feature pre-processing phase~\ref{SEC_feature_pre_processing}. The Cross-Domain Local Scaling (CSLS) metric is employed to measure the similarity and identify the most similar entities as candidates.
In the first iteration, the process only selects the top entity as the candidate.
Subsequently, the process gradually increases the number of potential entities (e.g., 1 to 10, then 20) in further iterations.
This iterative expansion approach aims to reduce the number of entities for comparison, enhancing the process's efficiency.

\subsubsection{Stage 2: Reasoning and Rethinking}
Utilizing the KG-Code translation module, this stage involves a detailed, multi-dimensional assessment of each candidate entity's alignment with the target entity in a dialogue form.

For reasoning, through the in-context learning along with few-shot cases, the model computes alignment scores based on name, structure, temporal, and generated entity description step by step. The detailed prompt can be found in Appendix~\ref{appendix_prompt}.

It then rethinks these collected results: if the top candidate's score significantly exceeds others and meets the confidence threshold, the alignment is considered satisfactory. Otherwise, the model revisits the candidate collecting results, expanding its search scope to reassess alignments with a broader candidate list.
This iterative refinement ensures comprehensive evaluation, significantly enhancing the final EA results' accuracy.

The two-stage EA strategy optimizes both the transparency of the EA process and performance while maintaining efficiency, in line with objective \textbf{O3}.
The detailed pseudo-code of this two-stage strategy is illustrated in Algorithm~\ref{ChatEA_algorithm}.

\begin{algorithm}[t]
\caption{Two-Stage EA Strategy in ChatEA}
\label{ChatEA_algorithm}
\footnotesize 
\begin{algorithmic}[1]
\State \textbf{Input:} The KG pair to be aligned \{$\mathcal{KG}_1$, $\mathcal{KG}_2$\}
\State \textbf{Output:} Aligned entity pairs $C$
\State features $\gets$ \Call{Feature Pre-processing}{$\mathcal{KG}_1, \mathcal{KG}_2$}
\For{scope $\gets$ $\{1, 10, 20\}$}
    \State \textit{// Stage 1: Candidate Collecting}
    \State cand $\gets$ \Call{Collect Candidates}{features, scope}
    \State  \textit{// Stage 2: Reasoning and Rethinking}
    \State align pair $\gets$ \Call{Reasoning}{cand, features}
    \State \textit{isSatisfied} $\gets$ \Call{Rethinking}{align pair}
    
    \If{\textit{isSatisfied}}
        \State Aligned entity pairs $C$ $\gets$ align pair
        \State \textbf{break}
    \EndIf
\EndFor
\State \Return Aligned entity pairs $C$
\end{algorithmic}
\end{algorithm}
\vspace{-5pt}









\section{Experiments}
In this section, we evaluate the ChatEA to ascertain its effectiveness in EA tasks.
Our investigation is guided by three pivotal research questions:

\begin{itemize}[leftmargin=*]
    \vspace{-2pt}
    \item \textbf{RQ1: Whether ChatEA overcomes the current EA limitations?} It delves into how ChatEA addresses the shortcomings of existing EA methods, aiming to validate its advancements.
    \vspace{-4pt}
    
    \item \textbf{RQ2: What is the effectiveness of ChatEA's each component?} This analysis focuses on evaluating the individual contributions and efficacy of ChatEA’s components.
    \vspace{-4pt}
    
    \item \textbf{RQ3: Does the ChatEA framework successfully balance accuracy and efficiency in EA?} This examination assesses whether ChatEA manages to strike an optimal balance between high accuracy and computational efficiency, a crucial aspect of its practical application.
    \vspace{-2pt}
\end{itemize}

\subsection{Experiment Settings}
Here, we introduce the datasets, baselines, model settings, and evaluation metrics in experiments.

\subsubsection{Datasets}
We conduct experiments on four entity alignment datasets. The statistics of these selected datasets are summarized in Table\ref{tb:dataset}. 

\textbf{DBP15K(EN-FR)} and \textbf{DBP-WIKI}~\cite{OpenEA} are two simple EA datasets, which share a similar structure for their KG pairs, with an equivalent number of entities. Furthermore, the structural features, such as the number of facts and density, of these two datasets closely align.
\textbf{ICEWS-WIKI} and \textbf{ICEWS-YAGO}~\cite{jiang2023rethinking} are two complex EA datasets~\footnote{https://github.com/IDEA-FinAI/Simple-HHEA}. Here, the KG pairs exhibit significant heterogeneity, differing not only in the number of entities but also in structural features. Notably, the quantity of anchors does not equal the number of entities. Consequently, aligning these complex datasets poses greater challenges.


\begin{table*}[t]
    \centering
    \setlength{\tabcolsep}{2pt}
    \begin{adjustbox}{max width=0.7\textwidth}
    \begin{tabular}{cc cccccc}
        \toprule
        \multicolumn{2}{c}{\bfseries Dataset}&\bfseries \#Entities&\bfseries \#Relations&\bfseries \#Facts&\bfseries Density&\bfseries \#Anchors&\bfseries Temporal\\
        \cmidrule(lr){1-2} \cmidrule(lr){3-8}
        \multirow{2}{*}{\textbf{DBP15K(EN-FR)}}&EN&15,000&193&96,318&6.421&\multirow{2}{*}{15,000}&No\\
        &FR&15,000&166&80,112&5.341&&No\\
        \midrule
        \multirow{2}{*}{\textbf{DBP-WIKI}}&DBP&100,000&413&293,990&2.940&\multirow{2}{*}{100,000}&No\\
        &WIKI&100,000&261&251,708&2.517&&No\\
        \midrule
        \multirow{2}{*}{\textbf{ICEWS-WIKI}}&ICEWS&11,047 &272 &3,527,881 &319.352 &\multirow{2}{*}{5,058} &Yes\\
        &WIKI&15,896 &226 &198,257 &12.472 &&Yes\\
        \midrule
        \multirow{2}{*}{\textbf{ICEWS-YAGO}} &ICEWS &26,863 &272 &4,192,555 &156.072 &\multirow{2}{*}{18,824} &Yes\\
        &YAGO &22,734 &41 &107,118 &4.712& &Yes\\
        \bottomrule
    \end{tabular}
    \end{adjustbox}
    \caption{The detailed statistics of the datasets. \textit{Temporal} denotes whether the dataset contains temporal information.}
    \label{tb:dataset}
\end{table*}

\begin{table*}[t!]
    \centering
    \setlength{\tabcolsep}{3pt}
    \begin{adjustbox}{max width=\textwidth}
        \begin{tabular}{lc|ccc|ccc|ccc|ccc}
            \toprule
            \multicolumn{2}{c|}{\textbf{Models}} &\multicolumn{3}{c|}{\textbf{DBP15K(EN-FR)}} &\multicolumn{3}{c|}{\textbf{DBP-WIKI}} &\multicolumn{3}{c|}{\textbf{ICEWS-WIKI}} &\multicolumn{3}{c}{\textbf{ICEWS-YAGO}}\cr
            &&Hits@1 &Hits@10 &MRR &Hits@1 &Hits@10 &MRR &Hits@1 &Hits@10 &MRR &Hits@1 &Hits@10 &MRR\cr
            \midrule
            &MTransE&0.247&0.577&0.360&0.281&0.520&0.363&0.021&0.158&0.068&0.012&0.084&0.040\cr
            &AlignE&0.481&0.824&0.599&0.566&0.827&0.655&0.057&0.261&0.122&0.019&0.118&0.055\cr
            &BootEA&0.653&0.874&0.731&0.748&0.898&0.801&0.072&0.275&0.139&0.020&0.120&0.056\cr
            &GCN-Align&0.411&0.772&0.530&0.494&0.756&0.590&0.046&0.184&0.093&0.017&0.085&0.038\cr
            &RDGCN&0.873&0.950&0.901&0.974&0.994&0.980&0.064&0.202&0.096&0.029&0.097&0.042\cr
            
            &Dual-AMN&0.954&0.994&0.970&0.983&0.996&0.991&0.083&0.281&0.145&0.031&0.144&0.068\cr
            
            &TEA-GNN&-&-&-&-&-&-&0.063&0.253&0.126&0.025&0.135&0.064\cr 
            
            &TREA&-&-&-&-&-&-&0.081&0.302&0.155&0.033&0.150&0.072\cr 
            
            &STEA&-&-&-&-&-&-&0.079&0.292&0.152&0.033&0.147&0.073\cr 
            &BERT&0.937&0.985&0.956&0.941&0.980&0.963&0.546&0.687&0.596&0.749&0.845&0.784\cr
            &FuAlign&0.936&0.988&0.955&0.980&0.991&0.986&0.257&0.570&0.361&0.326&0.604&0.423\cr
            
            &BERT-INT&\underline{0.990}&\underline{0.997}&\underline{0.993}&\textbf{0.996}&\underline{0.997}&\underline{0.996}&0.561&0.700&0.607&0.756&0.859&0.793\cr
            
            
            &Simple-HHEA&0.959&0.995&0.972&0.975&0.991&0.988&\underline{0.720}&\underline{0.872}&\underline{0.754}&\underline{0.847}&\underline{0.915}&\underline{0.870}\cr
            \midrule
            &ChatEA&\textbf{0.990}&\textbf{1.000}&\textbf{0.995}&\underline{0.995}&\textbf{1.000}&\textbf{0.998}&\textbf{0.880}&\textbf{0.945}&\textbf{0.912}&\textbf{0.935}&\textbf{0.955}&\textbf{0.944}\cr
            \bottomrule
    \end{tabular}
    \end{adjustbox}
    \caption{Main experiment results on the four datasets. \textit{Bold}: the best result; \textit{Underline}: the runner-up result.}
    \label{tb:main_results}
    \vspace{-10pt}
\end{table*}

\subsubsection{Baselines}
After carefully reviewing existing studies. We selected 11 state-of-the-art EA methods, which cover different input features, and KRL techniques.
These include translation-based methods such as MTransE~\cite{chen2017multilingual} AlignE~\cite{sun2018bootstrapping}, and BootEA~\cite{sun2018bootstrapping},
GNN-based methods like GCN-Align~\cite{wang2018cross}, RDGCN~\cite{wu2019relation}, TREA~\cite{xu2022time}, TEA-GNN~\cite{xu2021time}, STEA~\cite{STEA}, Dual-AMN~\cite{mao2021boosting},
and other methods like BERT-INT~\cite{devlin2018bert} and FuAlign~\cite{wang2023fualign}.
Additionally, it is pertinent to note the impracticality of directly employing LLMs for EA due to the input length constraints. For instance, fully inputting the ICEWS-WIKI and ICEWS-YAGO requires around 67,642k and 78,257k tokens, respectively. Such amounts surpass the usual LLM input limit (e.g., 128k tokens), making direct comparisons impractical.

\subsubsection{Model Configuration}
For LLM selection, we adopt the llama2-70b-chat~\cite{touvron2023llama} as our backbone, which is open-source and widely adopted. We also validate other representative LLMs in Section~\ref{SEC_llm_study_ChatEA}.

For consistency in the evaluation, the baseline models in the experiments adhere to the hyper-parameter specifications given in their original publications, except standardizing hidden dimensions to $d=64$ to ensure a fair comparison.

We followed the 3:7 splitting ratio in training/ testing data.
All models underwent identical preprocessing to prepare initial features for input.
Our development environment was PyTorch, and the experiments were conducted on a Ubuntu machine equipped with four 40GB NVIDIA A100 GPUs.

\subsubsection{Initial Feature Setup}
In our study, all EA models utilizing entity name information share the same entity name embeddings. Specifically, for {DBP15K(EN-FR)}, we obtain entity names using machine translation. For {DBP-WIKI}, we map QIDs into entity names. For {ICEWS-WIKI} and {ICEWS-YAGO}, we use the original entity names. After text feature extraction, we employ BERT with a whitening strategy~\cite{su2021whitening} to obtain the initial name embeddings.
Structure-based EA methods that do not utilize entity name information are initialized random initialization of embeddings according to their original method-specific configurations.

\subsubsection{Evaluation Metrics}
In line with widely adopted evaluation methods in EA research, we use two metrics for evaluation:
(1) {Hits@k}, measuring the percentage of correct alignments within the top $k$ ($k = 1, 10$) matches.
(2) {Mean Reciprocal Rank (MRR)}, reflecting the average inverse ranking of correct results. Higher values in Hits@$k$ and MRR indicate superior performance in the EA task.

\subsection{Main Experiment Results}
The comprehensive comparison conducted to address \textbf{RQ1} underscores ChatEA's consistent superiority or equivalence to state-of-the-art EA methods across various datasets, as highlighted in Table~\ref{tb:main_results}.

Specifically, ChatEA showcases remarkable performance, achieving a Hits@1 score of 0.990 on the DBP15K(EN-FR) dataset, equalling the performance of BERT-INT. On the DBP-WIKI dataset, it records a Hits@1 score of 0.995, slightly surpassed by BERT-INT's score of 0.996. The distinction becomes more pronounced on the ICEWS-WIKI and ICEWS-YAGO datasets, where ChatEA's Hits@1 scores of 0.880 and 0.935 respectively, which outperforms the best SOTA result (Simple-HHEA) by 16\% and 8.8\%, respectively.

This evidence leads to a critical examination of traditional EA methods. Especially GNN-based approaches reveal their limitations when faced with the highly heterogeneous KGs. The fundamental issue lies in their sole reliance on input KG data, which lacks the breadth of contextual information, and the constraints of KRL methods that are not equipped to handle such complexity.

As a comparison, our proposed ChatEA not only enriches entity descriptions with extensive background knowledge but also introduces an innovative two-stage EA strategy.
This approach significantly reduces dependency on input KG data and addresses the over-reliance on entity embedding comparisons. By utilizing LLMs' advanced reasoning capabilities, ChatEA refines alignment accuracy and effectively navigates the shortcomings of traditional KRL-based methods.


\subsection{Ablation Study}
To address \textbf{RQ2}, and assess the contribution of each component in ChatEA, we conducted ablation studies on the ICEWS-WIKI and ICEWS-YAGO. These studies aim to determine the individual benefits of components of ChatEA and investigate their influence on the base LLM's performance. The results are presented in Table~\ref{tb: ablation_study_ChatEA} and Table~\ref{tb: llm_study_ChatEA}.

\subsubsection{Effectiveness of Each Component}
To assess the impact of LLM, ChatEA (\textit{w/o} llm) excludes the two-stage EA strategy, relying solely on entity embeddings.
In comparison, the original ChatEA achieves significant performance gains (18.5\% and 12.5\% in Hits@1), highlighting the importance of both background knowledge and the reasoning ability of LLM in boosting EA accuracy.

ChatEA (\textit{w/o} name), ChatEA (\textit{w/o} structure), and ChatEA (\textit{w/o} temporal) respectively omit both the input data and relevant functions of name, structure, and temporal in the KG-Code translation module.
The results prove that the name, structure, and temporal information play a significant role in EA.

ChatEA (\textit{w/o} code) replaces the KG-code translation module by directly giving the entity name and tuples as LLM's input, resulting in a marked decrease in performance, which validates the effectiveness of the KG-Code translation in facilitating an understanding of KGs by the LLM.

ChatEA (\textit{w/o} desc) excludes entity descriptions in the two-stage EA strategy, leading to a performance decline. It reveals that generating entity descriptions based on the LLM’s background knowledge effectively activates contextual about entities stored in LLMs, which is crucial for accurate EA.


\begin{table}[t]
\centering
\setlength{\tabcolsep}{3pt}
\resizebox{0.9\linewidth}{!}{
\begin{tabular}{l|cc|cc}
    \toprule
    \multirow{2}{*}{\ \ \ \ \ \ \textbf{Settings}} &\multicolumn{2}{c}{\textbf{ICEWS-WIKI}} &\multicolumn{2}{c}{\textbf{ICEWS-YAGO}}\cr
    \cmidrule(lr){2-3}\cmidrule(lr){4-5}
    & Hits@1 & MRR & Hits@1 & MRR\cr
    \midrule
    \textbf{ChatEA} & \textbf{0.880} & \textbf{0.912} & \textbf{0.935} & \textbf{0.944} \cr
    \ - \textit{w/o} llm & 0.695 & 0.767 & 0.810 & 0.866 \cr
    \ - \textit{w/o} name & 0.640 & 0.709 & 0.685 & 0.747 \cr
    \ - \textit{w/o} structure & 0.860 & 0.891 & 0.925 & 0.937 \cr
    \ - \textit{w/o} temporal & 0.870 & 0.879 & 0.925 & 0.939 \cr
    \ - \textit{w/o} code & 0.810 & 0.831 & 0.870 & 0.883 \cr
    \ - \textit{w/o} description & 0.805 & 0.826 & 0.855 & 0.872 \cr
    \bottomrule
\end{tabular}}
\caption{Ablation study of ChatEA.}
\label{tb: ablation_study_ChatEA}
\end{table}

\begin{table}[t]
\vspace{-5pt}
\centering
\setlength{\tabcolsep}{3pt}
\resizebox{0.90\linewidth}{!}{
\begin{tabular}{l|cc|cc}
    \toprule
    \multirow{2}{*}{\ \ \ \ \ \ \textbf{Settings}} &\multicolumn{2}{c}{\textbf{ICEWS-WIKI}} &\multicolumn{2}{c}{\textbf{ICEWS-YAGO}}\cr
    \cmidrule(lr){2-3}\cmidrule(lr){4-5}
    & Hits@1 & MRR & Hits@1 & MRR\cr
    \midrule
    \textbf{ChatEA} & & & & \cr    
    \ - \textit{w/} llama2-70b & \underline{0.880} & \underline{0.912} & \underline{0.935} & \underline{0.944} \cr
    \ - \textit{w/} llama2-13b & 0.455 & 0.553 & 0.520 & 0.595 \cr
    \ - \textit{w/} gpt-3.5 & 0.860 & 0.895 & 0.875 & 0.913 \cr
    \ - \textit{w/} gpt-4 & \textbf{0.955} & \textbf{0.956} & \textbf{0.965} & \textbf{0.965} \cr
    \bottomrule
\end{tabular}}
\caption{ChatEA's performance with different LLMs.}
\label{tb: llm_study_ChatEA}
\vspace{-10pt}
\end{table}

\subsubsection{Performance with Different LLMs}~\label{SEC_llm_study_ChatEA}
Given ChatEA’s flexibility of plug-and-play for different LLMs, we evaluate how different backbone LLMs affect its performance on ICEWS-WIKI and ICEWS-YAGO, as shown in Table~\ref{tb: llm_study_ChatEA}.
The results show the enhanced capability of GPT-4 in boosting the performance of ChatEA rather than three other LLMs. 
Experiments with LLAMA2 at different scales (13b and 70b) highlight positive relationships between model size and ChatEA's efficacy.
As the capabilities of LLMs continue evolving, it is anticipated that ChatEA's proficiency in the EA task will correspondingly enhance.


\subsubsection{Influence of Entity Embeddings}~\label{SEC_entity_embedding}
We conduct entity embedding influence experiments in response to \textbf{RQ1} and \textbf{RQ2}. This experiment injects random noise into the dimensions of entity embeddings learned by KRL-based EA methods (i.e., Simple-HHEA) at ratios from 0\% to 80\%, simulating different qualities of entity embeddings. Subsequently, we contrast the performance of ChatEA with Simple-HHEA, which solely utilizes entity embedding comparison for EA.

\begin{figure}[t]
    \centering
    \includegraphics[width=1\columnwidth]{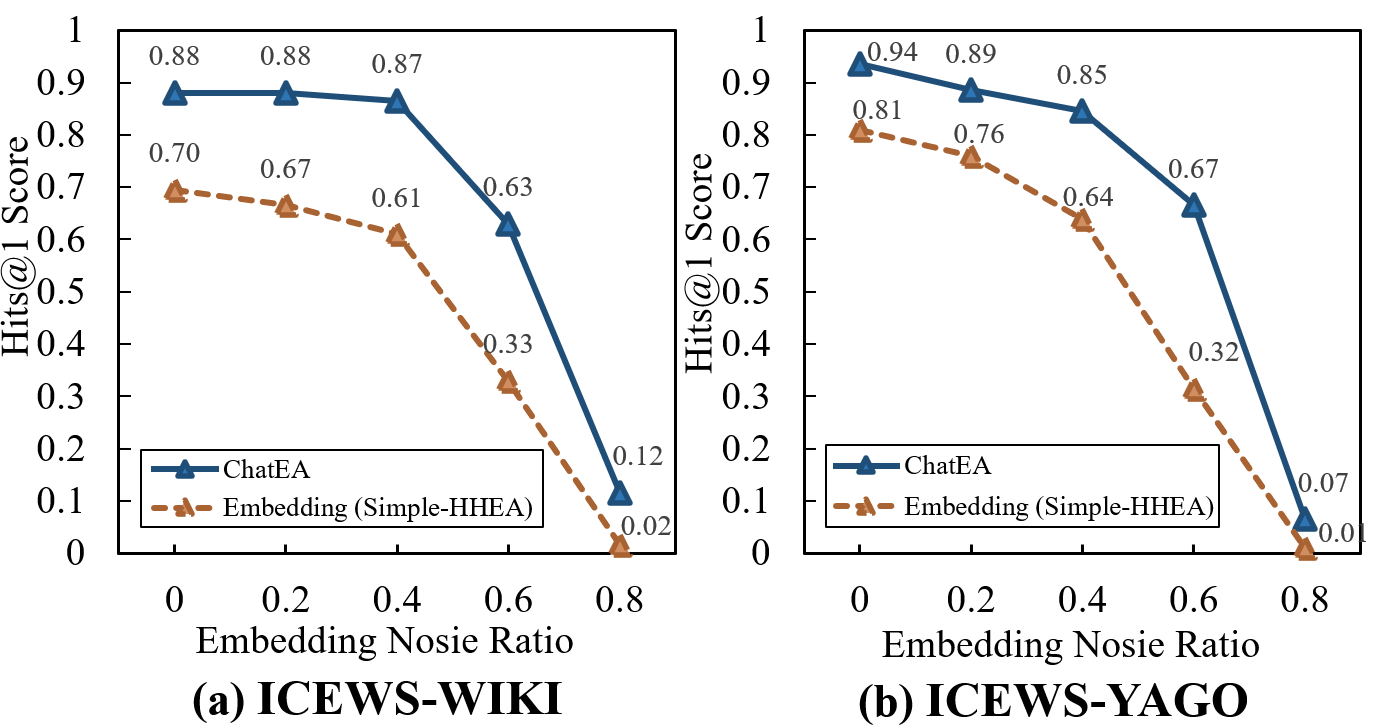}
    \caption{Performance influence of entity embeddings comparing ChatEA and Simple-HHEA for EA.}
    \label{fig:fig_embedding}
    \vspace{-15pt}
\end{figure}

The results, as shown in Figure~\ref{fig:fig_embedding},
indicate that when the noise ratio ranges from 0\% to 40\%, for KRL that achieves EA by directly comparing embeddings, the hits@1 results on ICEWS-WIKI and ICEWS-YAGO drop from 0.70 and 0.81 to 0.61 and 0.64, decreasing by 0.19 and 0.17, respectively. In contrast, the performance decrease for ChatEA is only 0.01 and 0.09, ensuring stable performance and robustness in EA. Remarkably, at 60\% noise, ChatEA still outperforms the single-embedding approach with 40\% noise, validating its efficacy.
This experiment demonstrates that even when the KRL is not good enough, ChatEA can also achieve stable performance.

Additionally, The candidate collection phase involves three rounds, considering up to 20 top candidates. A performance decline is observed when the embedding noise ratio exceeds 80\%, primarily because correct answers often do not appear among the top 20 candidates. As shown in the results, in the ICEWS-WIKI/YAGO datasets, with a noise ratio above 80\%, the ground truth entities fall outside the top 20 candidates in 62.9\% and 74.2\% of cases, respectively.
Therefore, expanding the candidate pool could potentially improve performance in scenarios with low-quality entity embeddings.

\subsection{Case Study}
To intuitively study the superiority of ChatEA, we illustrate a case chosen from the test sets of ICEWS-WIKI.
As shown in prompt case in Table~\ref{tb:detail_case_prompt} and output case in Table~\ref{tb:detail_case_output} in Appendix, ChatEA demonstrates its advanced capability by accurately aligning the \textit{British Monarch} entity with \textit{Monarchy\_of\_the\_United\_Kingdom}.
This success is achieved through an integrated approach that ChatEA strongly comprehends and reasoning with multiple attributes of entities across KGs, enhanced by the analytical prowess of LLM in the ChatEA. Thus, ChatEA can refine alignment results of entity embedding comparison through an explicit reasoning process.
Contrastingly, knowledge embedding methods incorrectly aligned \textit{British Monarch} with \textit{British\_Raj} (an Indian historical period related, not a political position), shows their limitations in external knowledge.

\subsection{Efficiency Analysis}
We conduct discussions about how ChatEA optimizes efficiency while maintaining accuracy.
In the two-stage EA strategy, ChatEA implements a three-round iteration rather than a single, intricately tailored to reasoning in complex datasets.
 
\begin{figure}[t]
    \centering
    \includegraphics[width=1\columnwidth]{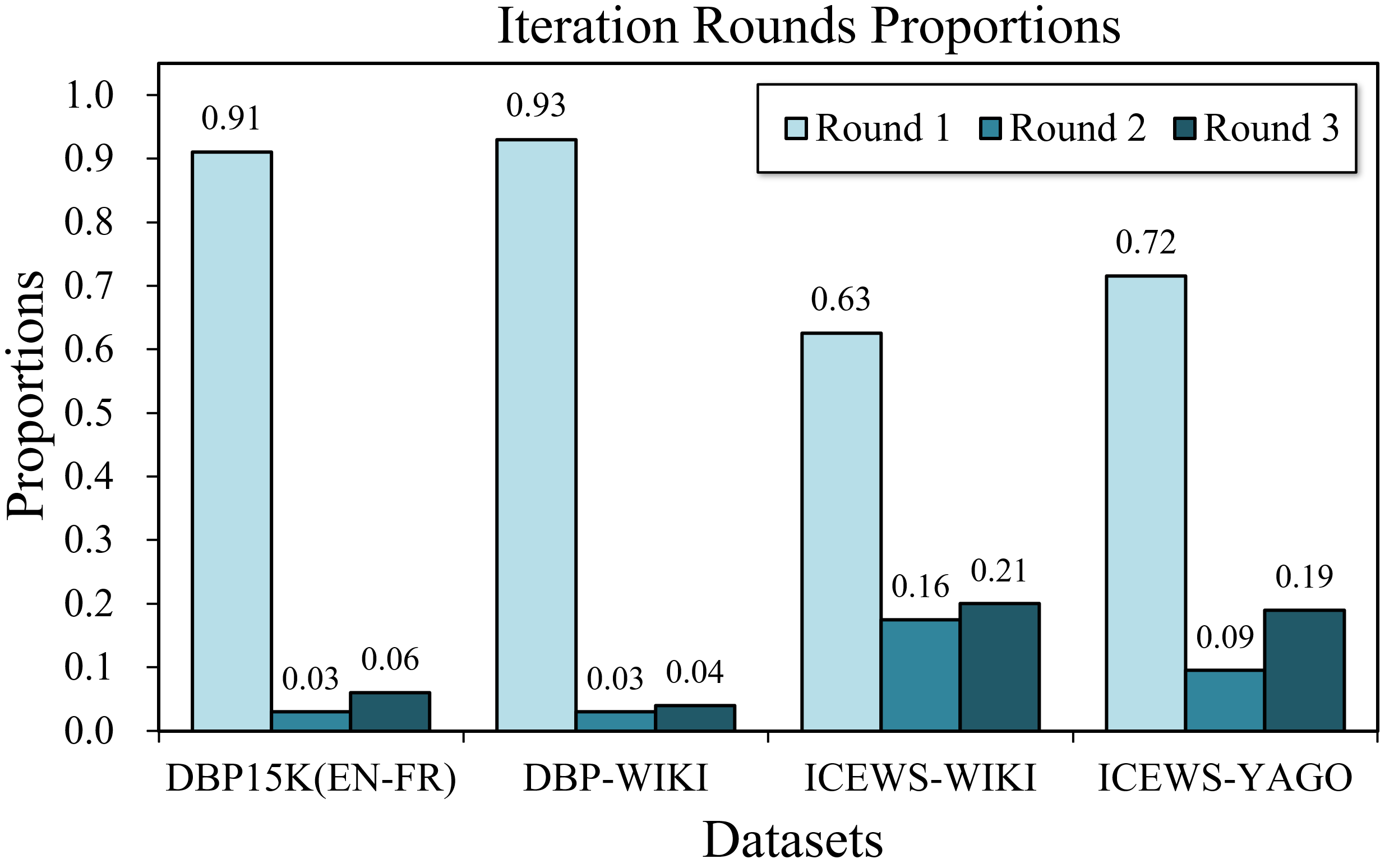}
    \caption{Proportions of iteration rounds of ChatEA's two-stage reasoning on different datasets.}
    \label{fig:fig_efficiency}
    \vspace{-5pt}
\end{figure}

As illustrated in Figure~\ref{fig:fig_efficiency}, with simpler datasets (i.e., DBP15K(EN-FR) and DBP-WIKI) where the entity feature preprocessing procedure is well-performing, ChatEA tends to converge faster, leading to better utilization of resources and higher efficiency.
Conversely, for more complex datasets(ICEWS-WIKI and ICEWS-YAGO), ChatEA inclines towards collecting more candidates and conducting thorough reasoning across extra iterations.
This adaptive methodology guarantees the maintenance of accuracy while optimizing LLM resource utilization, achieving a balance between accuracy and efficiency.
Additionally, the comparison between ChatEA and its variant(\textit{w/o} two-stage) in Table~\ref{tb: llm_efficiency} also demonstrates the superiority of the two-stage EA strategy in conserving resources and reducing time consumption.

From the perspective of application scenarios, ChatEA is designed for settings where high accuracy in EA is crucial, often prioritizing reliability over timeliness.
This is evident in experiments, where ChatEA shows superior accuracy, even if it compromises some efficiency.
Furthermore, ChatEA is adaptive to different LLMs, which positions it to benefit from advancements in LLM including not only accuracy but also efficiency.

\begin{table}[t]
\centering
\setlength{\tabcolsep}{3pt}
\resizebox{\linewidth}{!}{
\begin{tabular}{l|cc|cc}
    \toprule
    \multirow{2}{*}{\ \ \ \ \ \textbf{Settings}} &\multicolumn{2}{c}{\textbf{ICEWS-WIKI}} &\multicolumn{2}{c}{\textbf{ICEWS-YAGO}}\cr
    \cmidrule(lr){2-3}\cmidrule(lr){4-5}
    & {avg.tokens} & {avg.time} & {avg.tokens} & {avg.time}\cr
    \midrule
    \textbf{ChatEA} & & & & \cr    
    - \textit{w/} llama2-70b & 11,380 & 63.4 & 8,950 & 46.5 \cr
    - \textit{w/} llama2-13b & 47,007 & 150.1 & 44,907 & 135.8 \cr
    - \textit{w/} gpt-3.5 & 19,145 & 23.7 & 16,067 & 18.9 \cr
    - \textit{w/} gpt-4 & 9,803 & 131.8 & 6,593 & 90.8 \cr
    - \textit{w/o} two-stage & 56,059 & 312.5 & 58,404 & 303.2 \cr
    \bottomrule
\end{tabular}}
\caption{Efficiency analysis of ChatEA. \textit{avg.tokens} and \textit{avg.time} respectively denote the average tokens and time (seconds) cost of ChatEA per target entity}
\label{tb: llm_efficiency}
\vspace{-10pt}
\end{table}

\section{Conclusion}

In this paper, we focus on harnessing the capabilities of LLMs for EA, leading to the development of ChatEA. This innovative framework is tailored to address three pivotal challenges: (1) enhancing LLMs' ability to interpret and understand KGs, (2) leveraging the inherent knowledge within LLMs for more effective EA, and (3) improving the efficiency of LLMs in EA contexts. Our comprehensive experiments, conducted across four representative datasets, underscore ChatEA's superiority, particularly in applications requiring high precision in EA. These findings further illuminate the significant potential of LLMs in EA tasks for explorations.


\section{Limitations}
Despite the impressive accuracy achieved by ChatEA in EA with its innovative architecture and integration of LLMs.
It is essential to consider the limitations associated with resource consumption and efficiency of LLMs. Currently, ChatEA is particularly suited for applications where high precision in entity alignment is a critical requirement.
However, in scenarios that prioritize efficiency and can tolerate a certain degree of accuracy reduction, the current implementation of ChatEA faces constraints due to the inherent limitations in the inference speed of existing LLMs. This is a crucial factor to consider, especially in time-sensitive or resource-constrained environments.

In ChatEA, methodological enhancements aimed at improving efficiency were integrated, and a thorough discussion on this aspect is presented in the Efficiency Analysis section of the paper. These optimizations are crucial in striking a balance between accuracy and performance, yet there remains room for improvement, such as model distillation.

Besides, the performance constraints in smaller-scale models are also worth exploring in the future. While ChatEA excels with larger LLMs, its performance is notably constrained when applied to models with smaller parameter scales. Future iterations of ChatEA may need to incorporate techniques like sparse fine-tuning (SFT) to optimize performance without relying on large-scale models.


\section*{Acknowledgements}
Thanks to all reviewers, their reviews are important for this research.
This paper is funded by the NSFC (No.62172393), Major Public Welfare Project of Henan Province (No.201300311200).

\section{Ethics Statement}
To the best of our knowledge, this work does not involve any discrimination, social bias, or private data. All the datasets are constructed from open-source KGs such as Wikidata, YAGO, ICEWS, and DBpedia.
Therefore, we believe that our study complies with the ACL Ethics Policy.

\bibliography{custom}





\newpage
\appendix

\section{Appendix}\label{sec:appendix}

\subsection{Detailed Feature Pre-processing}\label{appendix_feature}
For entity name pre-processing, utilizing \textit{BERT}~\cite{devlin2018bert}, entity names are transformed into initial embeddings, further refined through a \emph{feature whitening transformation}~\cite{su2021whitening}. This process, combining BERT and whitening, effectively captures entity semantics. A linear transformation $W_{\mathcal{T}}$ finalizes the entity name embeddings $\{{\textbf{h}^{name}_n}\}^N_{n=1}$.

For entity time pre-processing, incorporating temporal information, the encoder leverages \textit{Time2Vec}~\cite{kazemi2019time2vec} for time representation. Entity times are encoded into binary vectors, with Time2Vec providing a learnable representation capturing time continuity and periodicity. The entity time embeddings $h^{time}$ are obtained via a linear transformation $W_{\mathcal{T}}$.

For entity structure, employing a biased random walk balancing BFS and DFS~\cite{wang2023fualign}, the encoder generates paths within KGs. The probability of selecting an entity is defined by its proximity to other entities. The structure of KGs is captured through the Skip-gram model $SkipGram(\cdot)$ and a linear transformation $W_\mathcal{D}$, generating entity embeddings $\{{{dw}_n}\}^N_{n=1}$.

Finally, multi-view embeddings are computed by concatenating the embeddings, expressed as:
$$\{{\textbf{h}^{mul}_n}\}^N_{n=1} = \{{[ \textbf{h}^{name}_n \otimes \textbf{h}^{time}_n\otimes \textbf{h}^{d w}_n]}\}^N_{n=1}.$$

We adopt \textit{Margin Ranking Loss} as the loss function for training, and \emph{Cross-domain Similarity Local Scaling (CSLS)}~\cite{conneau2017word} as the distance metric to measure similarities between entity embeddings.

\subsection{Comparison with Other PLM-based EA Methods}
In response to your concern about the comparison of ChatEA with other PLM methods: AttrGNN, SDEA, and TEA, we have conducted experiments and attached the experiment results as shown in Table~\ref{tab:plm_comparison}. The experimental results also validate the superiority of ChatEA.

\begin{table}[t]
\centering
\setlength{\tabcolsep}{3pt}
\resizebox{0.75\linewidth}{!}{
\begin{tabular}{l|cc|cc}
    \toprule
    \multirow{2}{*}{\textbf{Model}} &\multicolumn{2}{c}{\textbf{ICEWS-WIKI}} &\multicolumn{2}{c}{\textbf{ICEWS-YAGO}}\cr
    \cmidrule(lr){2-3}\cmidrule(lr){4-5}
    & Hits@1 & MRR & Hits@1 & MRR\cr
    \midrule
    AttrGNN & 0.047 & 0.093 & 0.015 & 0.044 \cr    
    SDEA    & 0.122 & 0.205 & 0.049 & 0.138 \cr
    TEA     & 0.610 & 0.718 & 0.657 & 0.740 \cr
    ChatEA  & \textbf{0.880} & \textbf{0.912} & \textbf{0.935} & \textbf{0.944} \cr
    \bottomrule
\end{tabular}}
\caption{Comparison of ChatEA with other PLM-based EA methods on ICEWS-WIKI/YAGO datasets.}
\label{tab:plm_comparison}
\end{table}

\subsection{Comparison of ChatEA and the Baseline on Other Classical EA Datasets}
For the experiment result of ChatEA on other classical datasets (i.e., datasets DBP15K(ZH-EN), DBP15K(JA-EN), and DBP-YAGO), to ensure the comprehensiveness of the methods compared in our paper, we have also included additional experiments here of ChatEA on above three datasets.
The results are shown in Table~\ref{tab:dbp_performance}.

\begin{table}[t]
\vspace{-5pt}
\centering
\setlength{\tabcolsep}{3pt}
\resizebox{1\linewidth}{!}{
\begin{tabular}{l|cc|cc|cc}
    \toprule
    \multirow{2}{*}{\textbf{Model}} &
    \multicolumn{2}{c|}{\textbf{DBP15K(ZH-EN)}} &
    \multicolumn{2}{c|}{\textbf{DBP15K(JA-EN)}} &
    \multicolumn{2}{c}{\textbf{DBP-YAGO}} \\
    \cmidrule(lr){2-3} \cmidrule(lr){4-5} \cmidrule(lr){6-7}
    & Hits@1 & MRR & Hits@1 & MRR & Hits@1 & MRR \\
    \midrule
    BERT-INT & 0.968 & 0.977 & 0.964 & 0.975 &\textbf{ 0.999} & \textbf{1.000} \\
    ChatEA   & \textbf{0.980} & \textbf{0.984} & \textbf{0.985} & \textbf{0.993} & 0.998 & 0.999 \\
    \bottomrule
\end{tabular}}
\caption{Comparison of ChatEA and BERT-INT on classical EA datasets: DBP15K(ZH-EN), DBP15K(JA-EN), and DBP-YAGO.}
\label{tab:dbp_performance}
\vspace{-10pt}
\end{table}

These experiments also confirm that ChatEA demonstrates superior performance on the datasets above compared to past embedding-based methods, validating our framework's effectiveness across various challenging scenarios.

\subsection{Detailed Prompt of ChatEA}\label{appendix_prompt}
In this section, we illustrate the prompts of ChatEA in Table \ref{tb:kg2code_prompt}, \ref{tb:reasoning_prompt} and \ref{tb:rethinking_prompt}.

\begin{table*}[h]
    \centering
    \begin{adjustbox}{max width=\textwidth}
    \begin{tabular}{p{\textwidth}}
        \toprule
        {\bfseries KG-Code Translation Prompt}\cr
        \midrule
        A Knowledge Graph Entity is defined as follows:\cr\cr
        Class Entity:\cr
        \qquad def \_\_init\_\_(self, name, id, tuples=[]):\cr
        \qquad\qquad self.entity\_name = name\cr
        \qquad\qquad self.entity\_id = id\cr
        \qquad\qquad self.tuples = tuples\cr
        \qquad def get\_description(self, LLM):\cr
        \qquad\qquad description = LLM(self.entity\_name, self.tuples)\cr
        \qquad\qquad return description\cr
        \qquad def get\_neighbors(self):\cr
        \qquad\qquad neighbors = set()\cr
        \qquad\qquad for head\_entity, \_, tail\_entity, \_, \_ in self.tuples:\cr
        \qquad\qquad\qquad if head\_entity == self.entity\_name:\cr
        \qquad\qquad\qquad\qquad neighbors.add(tail\_entity)\cr
        \qquad\qquad\qquad else:\cr
        \qquad\qquad\qquad\qquad neighbors.add(head\_entity)\cr
        \qquad\qquad return list(neighbors)\cr
        \qquad def get\_relation\_information(self):\cr
        \qquad\qquad relation\_info = []\cr
        \qquad\qquad for \_, relation, \_, \_, \_ in self.tuples:\cr
        \qquad\qquad\qquad relation\_info.append(relation)\cr
        \qquad\qquad return relation\_info\cr
        \qquad def get\_time\_information(self):\cr
        \qquad\qquad time\_info = []\cr
        \qquad\qquad for \_, \_, \_, start\_time, end\_time in self.tuples:\cr
        \qquad\qquad\qquad time\_info.append((start\_time, end\_time))\cr
        \qquad\qquad return time\_info\cr\cr
        You are a helpful assistant, helping me align or match entities of knowledge graphs according to name information (self.entity\_name), description information (get\_description()), structure information (self.tuples, get\_neighbors(), get\_relation\_information()), time information (get\_time\_information()), YOUR OWN KNOWLEDGE.\cr\cr
        Your reasoning process for entity alignment should strictly follow this case step by step:\cr\cr
        \{\{\quad\textit{reasoning case}\quad\}\}\cr\cr
        [Output Format]: [NAME SIMILARITY] = A out of 5, [PROBABILITY OF DESCRIPTION POINTING SAME ENTITY] = B out of 5, [STRUCTURE SIMILARITY] = C out of 5, [TIME SIMILARITY] = D out of 5. NOTICE, A,B,C,D are in range [1, 2, 3, 4, 5], which respectively means [VERY LOW], [LOW], [MEDIUM], [HIGH], [VERY HIGH]. NOTICE, you MUST strictly output like [Output Format].\cr
        \bottomrule
    \end{tabular}
    \end{adjustbox}
    \caption{KG-Code Translation Prompt, which is also the system prompt.}
    \label{tb:kg2code_prompt}
\end{table*}

\begin{table*}[h]
    \centering
    \begin{adjustbox}{max width=\textwidth}
    \begin{tabular}{p{\textwidth}}
        \toprule
        {\bfseries Reasoning Prompt}\cr
        \midrule
        Now given [Main Entity] l\_e = Entity( \{\{ \textit{Name, ID and Tuples} \}\} ), and [Candidate Entity] r\_e = Entity( \{\{ \textit{Name, ID and Tuples} \}\} ),\cr\cr
        - Do [Main Entity] and [Candidate Entity] align or match? Think of the answer STEP BY STEP with name, description, structure, time, YOUR OWN KNOWLEDGE:\cr\cr
        Step 1, think of [NAME SIMILARITY] = A out of 5, using self.entity\_name.\cr\cr
        Step 2, think of [PROBABILITY OF DESCRIPTION POINTING SAME ENTITY] = B out of 5, using get\_description() and YOUR OWN KNOWLEDGE.\cr\cr
        Step 3, think of [STRUCTURE SIMILARITY] = C out of 5, using self.tuples, get\_neighbors() and get\_relation\_information().\cr\cr
        Step 4, think of [TIME SIMILARITY] = D out of 5, using get\_time\_information().\cr\cr
        NOTICE, the information provided above is not sufficient, so use YOUR OWN KNOWLEDGE to complete them.\cr\cr
        Output answer strictly in format: [NAME SIMILARITY] = A out of 5, [PROBABILITY OF DESCRIPTION POINTING SAME ENTITY] = B out of 5, [STRUCTURE SIMILARITY] = C out of 5, [TIME SIMILARITY] = D out of 5.\cr
        \bottomrule
    \end{tabular}
    \end{adjustbox}
    \caption{Reasoning Prompt}
    \label{tb:reasoning_prompt}
\end{table*}

\begin{table*}[h]
    \centering
    \begin{adjustbox}{max width=\textwidth}
    \begin{tabular}{p{\textwidth}}
        \toprule
        {\bfseries Rethinking Prompt}\cr
        \midrule
        Now given the following entity alignments: \cr
        [Main Entity]: \{\{ \textit{Name} \}\} -> \{\{ \textit{Align Pairs} \}\}\cr\cr
        Please answer the question: Do these entity alignments are satisfactory enough ([YES] or [NO])?\cr\cr
        Answer [YES] if they are relatively satisfactory, which means the alignment score of the top-ranked candidate meet the threshold, and is far higher than others; otherwise, answer [NO] which means we must search other candidate entities to match with [Main Entity].\cr\cr
        NOTICE, Just answer [YES] or [NO]. Your reasoning process should follow [EXAMPLE]s:\cr\cr
        \{\{ \textit{Examples} \}\}\cr\cr
        Just directly answer [YES] or [NO], don't give other text.\cr
        \bottomrule
    \end{tabular}
    \end{adjustbox}
    \caption{Rethinking Prompt}
    \label{tb:rethinking_prompt}
\end{table*}

\subsection{Detailed Output of ChatEA about the Case Study-Input}\label{appendix_case_study}
The detailed prompt and output of ChatEA about the Case Study are illustrated in Table \ref{tb:detail_case_prompt} and \ref{tb:detail_case_output}.
\begin{table*}[h]
    \centering
    \begin{adjustbox}{max width=\textwidth}
    \begin{tabular}{p{\textwidth}}
        \toprule
        {\bfseries \#\#\# PROMPT about case study}\cr
        \midrule
        Now given [Main Entity] l\_e = Entity('British Monarch', '7497', 'The British Monarch is the head of the monarchy of the United Kingdom, currently held by Queen Elizabeth II, who has reigned since 1952 and has made various visits to countries such as the United States, South Korea, and Lithuania, among others, while also hosting visits from foreign leaders and dignitaries.', [(Ireland, Host a visit, British Monarch, 2011-03, 2011-03), (British Monarch, Host a visit, Elizabeth II, 2011-05, 2011-05), (British Monarch, Make a visit, United States, 2007-05, 2007-05), (British Monarch, Make a visit, South Korea, 1999-04, 1999-04), (Elizabeth II, Make a visit, British Monarch, 2011-05, 2011-05)]), \cr\cr
        and [Candidate Entity] r\_e = Entity('Monarchy\_of\_the\_United\_Kingdom', '23393', 'The Monarchy of the United Kingdom is the constitutional monarchy that serves as the head of state of the United Kingdom, with the monarch appointed by the Governor of Hong Kong and holding various roles such as the Lord Chancellor, Master of the Rolls, and Lord President of the Council.', [(Monarchy\_of\_the\_United\_Kingdom, country, United\_Kingdom, \textasciitilde, \textasciitilde), (Governor\_of\_Hong\_Kong, appointed by, Monarchy\_of\_the\_United\_Kingdom, \textasciitilde, \textasciitilde), (Monarchy\_of\_the\_United\_Kingdom, instance of, Constitutional\_monarchy, \textasciitilde, \textasciitilde), (Chancellor\_of\_the\_Duchy\_of\_Lancaster, appointed by, Monarchy\_of\_the\_United\_Kingdom, \textasciitilde, \textasciitilde), (Deputy\_Prime\_Minister\_of\_the\_United\_Kingdom, appointed by, Monarchy\_of\_the\_United\_Kingdom, \textasciitilde, \textasciitilde)]), \cr\cr
        - Compared with other Candidate Entities, do [Main Entity] and [Candidate Entity] align or match? Think of the answer STEP BY STEP with name, description, structure, time, YOUR OWN KNOWLEDGE: \cr
        Step 1, think of [NAME SIMILARITY] = A out of 5, using self.entity\_name. \cr
        Step 2, think of [PROBABILITY OF DESCRIPTION POINTING SAME ENTITY] = B out of 5, using self.entity\_description. \cr
        Step 3, think of [STRUCTURE SIMILARITY] = C out of 5, using self.tuples, get\_neighbors() and get\_relation\_information(). \cr
        Step 4, think of [TIME SIMILARITY] = D out of 5, using get\_time\_information(). \cr
        NOTICE, the information provided above is not sufficient, so use YOUR OWN KNOWLEDGE to complete them.\cr
        Output answer strictly in format: [NAME SIMILARITY] = A out of 5, [PROBABILITY OF DESCRIPTION POINTING SAME ENTITY] = B out of 5, [STRUCTURE SIMILARITY] = C out of 5, [TIME SIMILARITY] = D out of 5.\cr
        \bottomrule
    \end{tabular}
    \end{adjustbox}
    \caption{The detailed prompt of ChatEA about the Case Study}
    \label{tb:detail_case_prompt}
\end{table*}
\begin{table*}[h]
    \centering
    \begin{adjustbox}{max width=\textwidth}
    \begin{tabular}{p{\textwidth}}
        \toprule
        {\bfseries \#\#\# OUTPUT about case study}\cr
        \midrule
        \text{[NAME SIMILARITY] = 5 out of 5,}\cr
        \text{[PROBABILITY OF DESCRIPTION POINTING SAME ENTITY] = 5 out of 5,}\cr
        \text{[STRUCTURE SIMILARITY] = 5 out of 5,}\cr
        \text{[TIME SIMILARITY] = 5 out of 5.}\cr\cr
        Explanation:\cr\cr
        \text{Step 1: [NAME SIMILARITY] = 5 out of 5}\cr
        The names "British Monarch" and "Monarchy of the United Kingdom" are identical, indicating a high name similarity.\cr\cr
        \text{Step 2: [PROBABILITY OF DESCRIPTION POINTING SAME ENTITY] = 5 out of 5}\cr
        The descriptions of both entities mention the same entity, "the United Kingdom," and describe similar roles and responsibilities, indicating a high probability of pointing to the same entity.\cr\cr
        \text{Step 3: [STRUCTURE SIMILARITY] = 5 out of 5}\cr
        Both entities have the same structure, as they are both instances of the Constitutional monarchy, with the monarch appointed by the Governor of Hong Kong and holding various roles such as the Lord Chancellor, Master of the Rolls, and Lord President of the Council.\cr\cr
        \text{Step 4: [TIME SIMILARITY] = 5 out of 5}\cr
        Both entities have the same time information, as they both refer to the current monarch, Queen Elizabeth II, who has reigned since 1952 and has made various visits to countries such as the United States, South Korea, and Lithuania, among others, while also hosting visits from foreign leaders and dignitaries.\cr\cr
        Therefore, the alignment score between the two entities is (5, 5, 5, 5), indicating a perfect match.\cr
        \bottomrule
    \end{tabular}
    \end{adjustbox}
    \caption{The detailed output of ChatEA about the Case Study-Output}
    \label{tb:detail_case_output}
\end{table*}

\subsection{Problematic Outputs of ChatEA with llama2-13b}\label{appendix_case_llama13b}
We illustrate two problematic outputs of ChatEA with llama2-13b in Table \ref{tb:promblem_output_13b_1} and \ref{tb:promblem_output_13b_2}. Due to space limitations, we only provide the name and description information of entities in the table.

In the first problematic output in Table \ref{tb:promblem_output_13b_1}, based on the names, descriptions, and other information of entities "Myanmar" and "Burundi", llama2-13b is able to make correct reasoning about the [NAME SIMILARITY], [PROBABILITY OF DESCRIPTION POINTING SAME ENTITY] and so on. However, llama2-13b does not correctly output scores in the specified format, but instead repeats the content of the prompt, for example, it outputs "Step1. Think of [NAME SIMILARITY] = A out of 5.", which is exactly the same as the prompt. As a result, ChatEA cannot extract similarity scores from the output of llama2-13b to calculate the correct final alignment score.

In the first problematic output in Table \ref{tb:promblem_output_13b_2}, llama2-13b cannot correctly identify entity information, leading to a large number of errors in the reasoning process. For example, llama2-13b think that the names of the entity "United Nations" and "Barack\_Obama" both contain "United" and "Nations", so it thinks that the name similarity is high and output "[NAME SIMILARITY]= 4 out of 5". But in fact the name of the entity "Barack\_Obama" does not contain "United" and "Nations". Another example is that, llama2-13b output "Both entities have similar descriptions, as
they are both international organizations", but in fact the description of entity "Barack\_Obama" indicates that entity "Barack\_Obama" is the President of the United States. Such reasoning errors prevent ChatEA from correctly calculating the alignment scores.

For LLMs with relatively small parameters such as llama2-13b, problems including information recognition errors, reasoning errors, output format errors and so on still exist. In future work, supervised fine-tuning is needed to activate the potential of these LLMs with relatively small parameters on this entity alignment task.

\begin{table*}[h]
    \centering
    \begin{adjustbox}{max width=\textwidth}
    \begin{tabular}{p{\textwidth}}
        \toprule
        {\bfseries \#\#\# Entity 1:}\cr
        \midrule
        Name = "Myanmar",\cr
        Description = "Myanmar is a country located in Southeast Asia, formerly known as Burma, with a diverse culture and a complex history of political and economic changes, bordering several countries including China, Laos, Thailand, and Bangladesh, and having diplomatic relations with many nations."\cr
        \midrule
        {\bfseries \#\#\# Entity 2:}\cr
        \midrule
        Name = "Burundi",\cr
        Description = "Burundi is a country located in the African Great Lakes region, bordered by the Democratic Republic of the Congo, Rwanda, Tanzania, and Lake Tanganyika, with a diverse cultural heritage and a history of political instability."\cr
        \midrule
        {\bfseries \#\#\# OUTPUT}\cr
        \midrule
        Sure, I can help you with that! Based on the information provided, here are my answers:\cr\cr
        Step 1: Think of [NAME SIMILARITY] = A out of 5.\cr
        Myanmar and Burundi have similar names, but they are not exactly the same. Myanmar is a country located in Southeast Asia, while Burundi is a country located in the African Great Lakes region.\cr\cr
        Step 2: Think of [PROBABILITY OF DESCRIPTION POINTING SAME ENTITY] = B out of 5.\cr
        The descriptions of Myanmar and Burundi are not identical, but they do share some similarities. Both descriptions mention the country's location and its diverse population. However, Myanmar's description also includes information about its complex history of political and social change, while Burundi's description focuses on its cultural heritage and history of political instability.\cr\cr
        Step 3: Think of [STRUCTURE SIMILARITY] = C out of 5.\cr
        Myanmar and Burundi have different structures in terms of their tuples and relation information. Myanmar has more tuples and a more complex structure, with information about its expressing intent to cooperate, hosting visits, and accusing other entities. Burundi, on the other hand, has fewer tuples and a simpler structure, with information about its membership in organizations and its part in a larger entity.\cr\cr
        Step 4: Think of [TIME SIMILARITY] = D out of 5.\cr
        Myanmar and Burundi have different time information. Myanmar's tuples include information about events that took place in 2007 and 2013, while Burundi's tuples include information about events that took place in 1963 and 2018.\cr\cr
        Therefore, based on the information provided, I would say that [Main Entity] Myanmar and [Candidate Entity] Burundi do not align or match perfectly. The name similarity is moderate, the probability of description pointing to the same entity is low, the structure similarity is low, and the time similarity is low.\cr
        \bottomrule
    \end{tabular}
    \end{adjustbox}
    \caption{One problematic output of ChatEA with llama2-13b, where the reasoning process is correct but the output format is incorrect.}
    \label{tb:promblem_output_13b_1}
\end{table*}
\begin{table*}[h]
    \centering
    \begin{adjustbox}{max width=\textwidth}
    \begin{tabular}{p{\textwidth}}
        \toprule
        {\bfseries \#\#\# Entity 1:}\cr
        \midrule
        Name = "United Nations",\cr
        Description = "The United Nations is an international organization that aims to promote peace, security, and cooperation among its member states, and to address global challenges such as poverty, inequality, and climate change."\cr
        \midrule
        {\bfseries \#\#\# Entity 2:}\cr
        \midrule
        Name = "Barack\_Obama",\cr
        Description = "Barack Obama was the 44th President of the United States, serving two terms from 2009 to 2017, and is the first African American to hold the office."\cr
        \midrule
        {\bfseries \#\#\# OUTPUT}\cr
        \midrule
        \text{[NAME SIMILARITY] = 4 out of 5,}\cr
        \text{[PROBABILITY OF DESCRIPTION POINTING SAME ENTITY] = 3 out of 5,}\cr
        \text{[STRUCTURE SIMILARITY] = 4 out of 5,}\cr
        \text{[TIME SIMILARITY] = 2 out of 5.}\cr\cr
        Explanation:\cr\cr
        1. Name similarity: Both entities have "United" in their names, and "Nations" in the second part of their names. Therefore, the name similarity is high.\cr\cr
        2. Probability of description pointing to same entity: Both entities have similar descriptions, as they are both international organizations. However, the description of the United Nations is more detailed and comprehensive, while the description of Barack Obama is more specific to his role as the 44th President of the United States. Therefore, the probability of description pointing same entity is moderate.\cr\cr
        3. Structure similarity: Both entities have a similar structure, with several tuples pointing to them. However, the United Nations has more tuples pointing to it, and the tuples are more diverse, including appeals, expressions of intent, consultations, and membership. Therefore, the structure similarity is high.\cr\cr
        4. Time similarity: Both entities have a similar time dimension, as they are both contemporary entities. However, the United Nations has a longer time dimension, as it was established in 1945, while Barack Obama's tenure as President was from 2009 to 2017. Therefore, the time similarity is moderate.\cr\cr
        Overall, the United Nations and Barack Obama align moderately well in terms of name, description, structure, and time, but not perfectly.\cr
        \bottomrule
    \end{tabular}
    \end{adjustbox}
    \caption{One problematic output of ChatEA with llama2-13b, where the reasoning process is incorrect.}
    \label{tb:promblem_output_13b_2}
\end{table*}
\end{document}